# Intent Expression Using Eye Robot for Mascot Robot System

Yoichi Yamazaki, Fangyan Dong, Yuta Masuda, Yukiko Uehara, Petar Kormushev,
Hai An Vu, Phuc Quang Le, and Kaoru Hirota
Department of Computational Intelligence and Systems Science
Tokyo Institute of Technology, Japan,  {yama, tou, hirota}@hrt.dis.titech.ac.jp

*Abstract*—An intent expression system using eye robots is proposed for a mascot robot system from a viewpoint of humatronics. The eye robot aims at providing a basic interface method for an information terminal robot system. To achieve better understanding of the displayed information, the importance and the degree of certainty of the information should be communicated along with the main content. The proposed intent expression system aims at conveying this additional information using the eye robot system. Eye motions are represented as the states in a pleasure-arousal space model. Changes in the model state are calculated by fuzzy inference according to the importance and degree of certainty of the displayed information. These changes influence the arousal-sleep coordinates in the space that corresponds to levels of liveliness during communication. The eye robot provides a basic interface for the mascot robot system that is easy to be understood as an information terminal for home environments in a humatronics society.

## I. INTRODUCTION

As technology advances, robots are expected to become widespread in home environments. Robots can provide an easily understandable interface for an information terminal, which doesn't require special input devices such as a keyboard or a mouse[1][2][3][4]. Some necessary elements for a domestic information appliances robot are IT (Information Technology), recognition technology, and emotion expression technology from a humatronics point of view. Several new results have already been achieved regarding IT and recognition technology.  Emotion expression technology is necessary to allow humans to understand robots[5][6][7][8][9]. Emotion expression technology makes humans and robots more comfortable and friendly when communicating with each other. Mentality expression is an essential component for friendly robot communication, and eye motions are suitable for the subtle expression of emotions[10]. A mentality expression system using an eye robot has already been proposed [11][12] for communication between human beings and robots.

The expression of emotions is based on an affinity pleasure-arousal space, where mentality status is calculated by fuzzy inference from a speech understanding module. The constructed eye robot expresses mentality by easily comprehensible eye gestures, and constitutes an interface for a new type of user friendly information terminal, the so called mascot robot system[11]. The mascot robot system consists of speech recognition modules, information recommendation modules, and the eye robots, that are integrated with RT (Robot Technology) Middleware developed by AIST Japan. Its development is part of an ongoing project "Development Project for a Common Basis of Next-Generation Robots" led by the NEDO organization (New Energy and Industrial Technology Development Organization).

In addition to the mentality expression system, an intent expression system (from a humatronics point of view) using an eye robot is proposed for the mascot robot system. The eye robot aims at providing a basic interface for an information terminal robot system. To achieve better understanding of the displayed information, the importance and the certainty of the information should be communicated along with the main content. The proposed intent expression system aims at conveying this additional information using the eye robot. The eye motions are represented as the states in a pleasure-arousal space model. Changes in the model state are calculated by fuzzy inference according to the importance and degree of certainty of the displayed information. These changes influence the arousal-sleep coordinates in the space which corresponds to levels of liveliness during communication. The eye robot provides a basic interface for the mascot robot system which is easily understandable as an information terminal for home environments in a humatronics based society.

An overview of the mascot robot system is mentioned in II. A fuzzy inference method for intent expression is presented in III. Experimental results using the eye robot are presented in IV.

This work was supported by Development Project for a Common Basis of Next-Generation Robots (sponsored by NEDO, Japanese government).

## II. MASCOT ROBOT SYSTEM

IT society is the society where everyone can enjoy the benefits of IT at any time and any place. Home robots are expected to function as a convenient and friendly interface for IT systems in the home environment. In this paper, a mascot robot system is proposed as an easily visible information terminal. Three components, i.e., an information proposal module, which offers information and choices that users may select from, a speech recognition module, which can easily be used by anyone, and five friendly eye robots, are integrated into a mascot robot system with the aid of RT Middleware developed by AIST Japan. The system's architecture is shown in Fig.1.

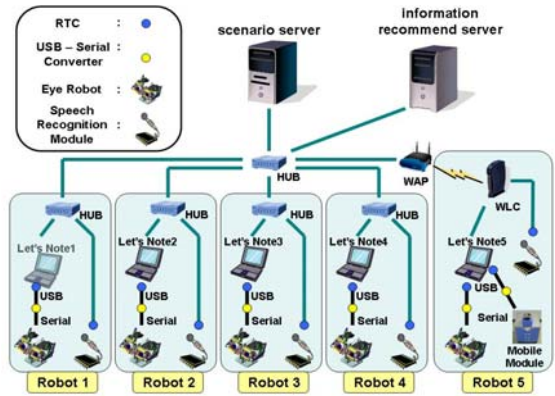

Fig. 1  Architecture of the mascot robot system

The information recommendation module collects information about users from their speech, and proposes interesting information for users [11]. The speech recognition module is developed for general robot systems by NEC Corporation in the Development Project for a Common Basis of Next-Generation Robots.

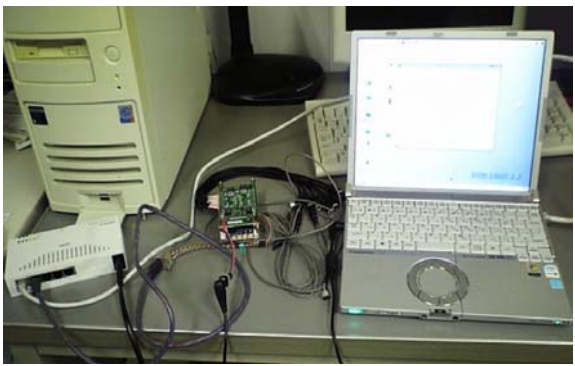

Fig. 2  Speech recognition module

This module in Fig.2 is compact and has high performance, e.g. it has a small size, low cost, low power consumption, and an audio input. The eye robot with mentality expression has been already proposed for communication between humans and robots. The expression of emotions is based on an affinity pleasure-arousal space, where mentality status is calculated by fuzzy inference from a speech understanding module. The developed eye robot expresses mentality by easily comprehensible eye gestures, and constitutes the friendly mascot robot system [11]. Four eye robots are stationary, and one eye robot is put on a mobile robot in Fig.3,4,5.

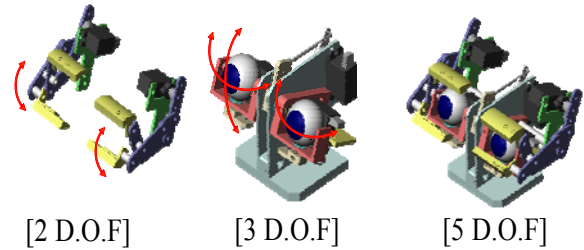

[2 D.O.F]    [3 D.O.F]    [5 D.O.F]

Fig. 3  Structure of the eye robot

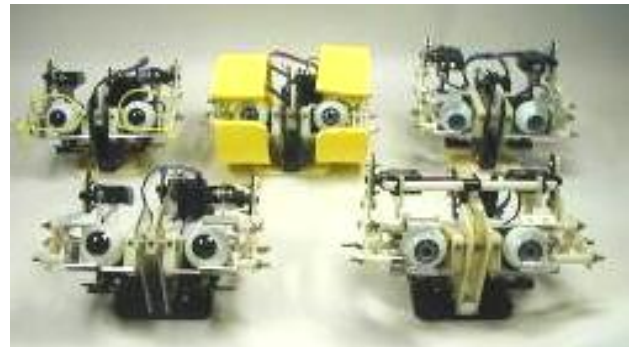

Fig. 4  Five eye robots for intent expression

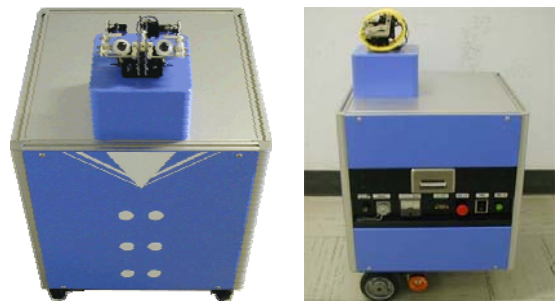

Fig. 5  Mobile type eye robot

The ability to move allows the robot to move closer to people since the speech recognition module has to be close to the source of the sound to function properly. These are all integrated with RT Middleware as shown in Fig.6, which has been developed by the National Institute of Advanced Industrial Science and Technology (AIST), with the aim of building a common basic framework for the robot industry. Its development is part of an ongoing project "Development Project for a Common Basis of Next-Generation Robots" led by the NEDO organization (New Energy and Industrial Technology Development Organization).

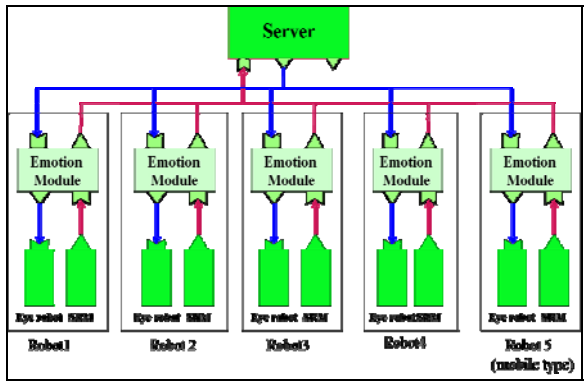

Fig. 6 System construction of five robots based on RTM

III. EYE ROBOT BASED INTENT EXPRESSION SYSTEM

A. *Eye Robot*

The Eye robot that can express eye motions is developed based on the mechanism of human eyes. Eye motion consists of the combination of eyelid motion and ocular motion. The structures of eyelid part and ocular part are shown in Fig. 4 . The eye robot has 2 degrees of freedom (D.O.F) for the eyelid part and 3 D.O.F for the ocular part. Both left and right eyelids have 1 D.O.F for opening, closing, and blinking. This eyelid part has a palpebra superior and a palpebra inferior. These two palpebras are linked to each other and have 1 D.O.F. Hence, the eyelid part as a whole has 2 D.O.F. The 3 D.O.F are given to binocular and ocular parts. The eye robot has 5 degrees of freedom, meaning that a smaller number of actuators is needed than that necessary for a facial expression system. Therefore, this system is appropriate for multi arranged type robot applications, especially from a cost point of view.

B. *Intent Expression by Eye Robot System*

A mentality expression system using the eye robot is proposed in [11]. Its input is language category information that is generated by a speech recognition system and its output is the expression of mentality using motions of the robot's eyes. The mechanical part of the eye robot which can produce eye motions is based on the mechanisms of the human eye for familiarity considerations. The eye robot has an eyelid part and an ocular part. The modality of expression with eyelids and the ocular part is based on a two-dimensional pleasure-arousal plane. The two-dimensional plane has a pleasure-displeasure axis and arousal-sleep axis. The pleasure-displeasure axis depends on the approval of the interlocutor, which is determined from the speech recognition module. Arousal-sleeping axis relates to liveliness during communication. To take into consideration the continuous transition of mentality during interlocution, the three-dimensional affinity pleasure-arousal space is proposed as an extension of the pleasure-arousal plane. The motivation for expression is determined by fuzzy inference in the affinity pleasure-arousal space (Fig.7).

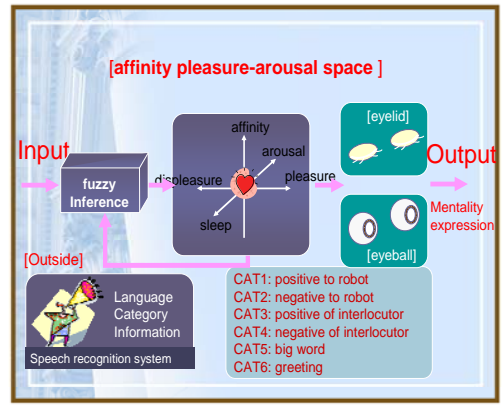

Fig.7 Affinity pleasure-arousal space

But, it is not enough for a robot application communicating with humans to just show the user the necessary information through the display unit and speech recognition parts could not recognize inputs using only one feature. Information display parts have an order of precedence when showing information. Therefore systems must be able to show information comprehensibly in addition to the main content. Human beings use nonverbal messages to support verbal information, and communicate with each other effectively. Their are mainly four types of nonverbal message, i.e. emotional messages that show mentality state of sender, emotional messages that show agreement or disagreement with verbal information, meta-communication messages that alter speech timing, and specific gestures. Robots need to use these methods just like people to be able to carry out friendly communication.

Eyes are important for communication, and their mechanisms are simple to build. The mentality expression system using an eye robot has been already proposed for communication between human beings and robots, and has been successful at offering a friendly atmosphere. There has been, however, no method to show emotional messages that express agreement or disagreement between verbal information and eye motion.

C. *Intent Expression for the Mascot Robot System*

In this paper, an intention expression system using the eye robot is proposed. This makes it possible for the robot agent to communicate non-verbally, and send emotional messages such as agreement, recommendation or sympathy (cf. Fig.8).

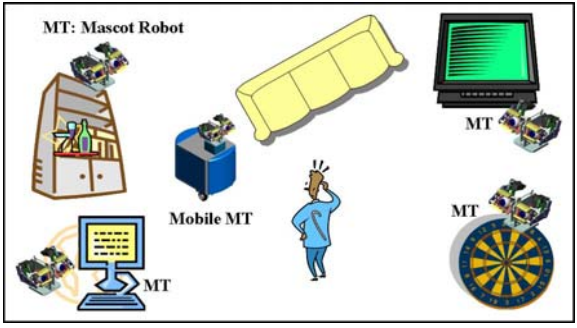

Fig. 8 The home environment for the mascot robot system

The system is designed for the mascot robot system. The intent expression system gives the mascot robot the ability to express the priority of recommendations according to the information recommendation module. It also makes it possible to express the degree of recommendation of the various items the robot proposes.

The speech recognition module of the mascot robot system doesn't output the level of confidence. Therefore in this proposed intention expression system, information from the recommendation module will be used, but potential information from the speech recognition module will not.

Mentality expression with the eye is independently assigned in the affinity pleasure-arousal space. The pleasure-arousal space consists of an arousal-sleep axis for the display of liveliness and the pleasure-displeasure axis for the display of real-time friendliness. Taking into consideration the variation of the mentality state during communication, the addition of an affinity axis to create an affinity pleasure-arousal space. The input of the intent expression system is the priority level of the information recommendation coming from the recommendation module and the mentality state as given by current position within the pleasure-arousal space. The output of the intent expression system is a movement of intent expression determined from the mentality state. Fuzzy inference is used for generating intent expressing movement based on the information recommendation priority level and the mentality state.

Input is represented by coordinates in the pleasure-arousal plane **S** and a priority of recommendation *r*. The *S* is expressed as

$$\begin{aligned} &S \in \mathbf{S} \\ &S(x_{pl}, x_{ar}), \\ &-200 \le x_{pl} \le 200, \\ &-200 \le x_{ar} \le 200. \end{aligned} \quad (1)$$

The output is $\Delta S$ a variation of $S$. $\Delta S$ is expressed as

$$\begin{aligned} &\Delta S(x_{pl}, x_{ar}) = (\Delta x_{pl}, \Delta x_{ar}), \\ &-50 \le \Delta x_{pl} \le 50, \\ &-50 \le \Delta x_{ar} \le 50 \quad . \end{aligned} \quad (2)$$

To obtain the fuzzy quantization, membership functions are defined for each element of input. To obtain the output, the center of area defuzzification method is used.

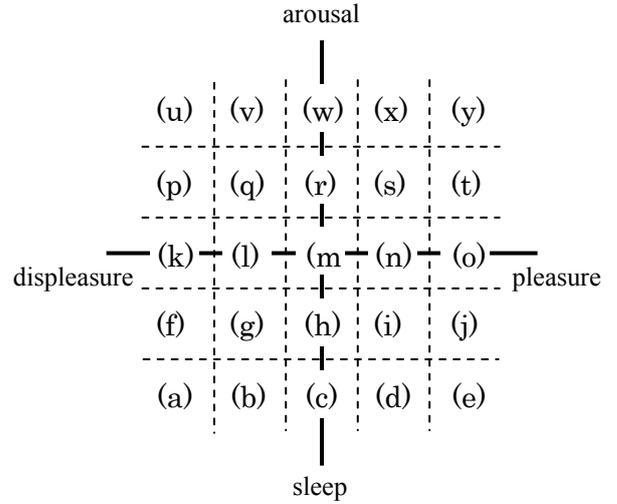

Fig. 9 Pleasure arousal plane

## IV. THE EXPERIMENTS FOR INTENT EXPRESSION WITH EYE ROBOT

### A. Experiments environment for intent expression

In the intent expression system using the eye robot, intent is expressed based on the level of recommendation generated by the information recommendation module from input information.

To determine the relationship between the expression movement and the recommendation level, evaluation experiments were performed. These experiments were based on the results of human evaluations using information collected via questionnaire.

Recommending or not recommending a book is used as a benchmark to determine 6 evaluation levels of recommendation degree Fig.10.

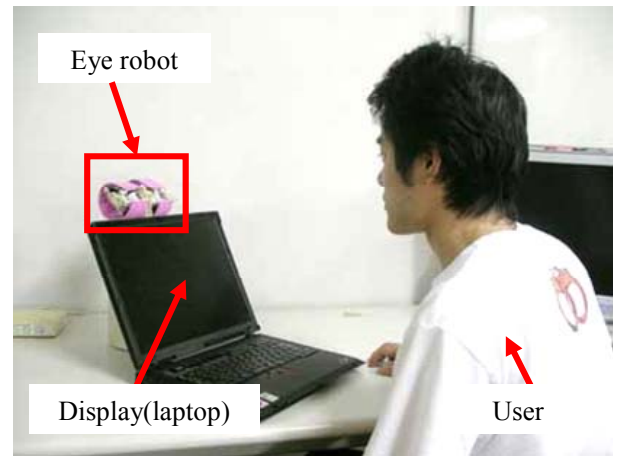

Fig. 10 Benchmark test

The experimental procedure is described in the followings.

1) The experiment's context (recommending a book) and the questionnaire are explained to the subject.

2) A recommended book is displayed, and then the robot expresses through movement 1 out 20 possible mentality states determined from pleasure-arousal space. The order of the movements is random.

3) The subject evaluates the expression movements and their variations using 6 grades.

Steps 2), 3) are repeated for all 20 possible mentality states..

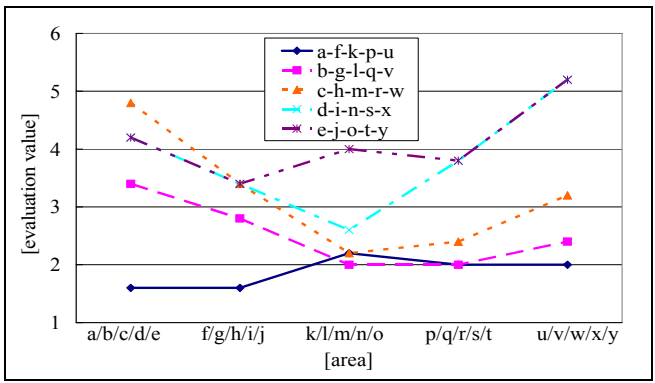

Fig. 11 Evaluation test 1

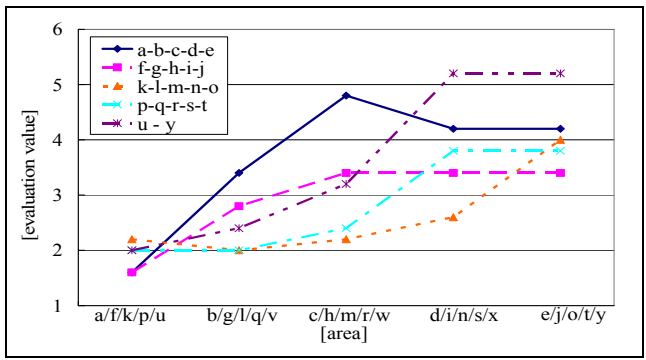

Fig. 12 Evaluation test 2

## V. CONCLUSIONS

A mascot robot system is proposed for functioning in home environments and coexisting with humans, and an intent expression system is demonstrated. The mascot robot system consists of 5 eye robots, a speech recognition module, and an information recommendation module which are all integrated using RT middleware technology.

The eye robot is based on human eyes of 5 years old boy, and has eyelids with 2 degrees of freedom and eyeballs with 3 degrees of freedom. Four of the eye robots are fixed and one is put on a mobile robot. For the better human and robot communications, an intent expression system is proposed.

This system permits the communication of non-verbal information along with the main content of the desired message. The system takes as input information the degree of confidence, outputs a degree of trust and calculates importance levels. These are combined to form a pleasure-arousal space, which along with providing a methods for expressing mentality state, also makes it possible to express intent.

The proposed system provides a user friendly interface so that humans and robots communicate in natural fashion with needing to use keyboard type input devices. This is accomplished through a speech recognition module. Compared to current information terminals, the proposed interface is user friendly and appealing to humans, so the mascot robot system using eye robots is suitable for wide spread family use.